\documentclass[format=acmsmall, nonacm, screen=True]{acmart}
\setcopyright{none}
\AtBeginDocument{%
  \providecommand\BibTeX{{%
    \normalfont B\kern-0.5em{\scshape i\kern-0.25em b}\kern-0.8em\TeX}}}

\def\*#1{\mathbf{#1}}
\def\~#1{\boldsymbol{#1}}

\usepackage{amsmath}
\usepackage{algorithm}
\usepackage{algpseudocode}
\usepackage{graphicx}
\usepackage{caption}
\usepackage{subcaption}

\setcopyright{acmcopyright}
\copyrightyear{2018}
\acmYear{2018}
\acmDOI{XXXXXXX.XXXXXXX}

\acmConference[Conference acronym 'XX]{Make sure to enter the correct
  conference title from your rights confirmation email}{June 03--05,
  2018}{Woodstock, NY}
\acmBooktitle{Woodstock '18: ACM Symposium on Neural Gaze Detection,
 June 03--05, 2018, Woodstock, NY} 
\acmPrice{15.00}
\acmISBN{978-1-4503-XXXX-X/18/06}

\usepackage{tikz}
\usepackage{threeparttable}
\usepackage{array}
\usepackage{geometry}
\usepackage[utf8]{inputenc}
\usepackage{graphicx} 
\usepackage{pgfplots}
\usepackage{caption}
\usepackage{graphicx}
\usepackage{bbm}
\usepackage{threeparttable}
\usepackage{soul}
\usepackage[misc]{ifsym}
\usepackage[capitalize]{cleveref}
\usepackage{rotating}
\usepackage{float}
\usepackage{longtable}
\usepackage{array}
\usepackage{amsmath}
\usepackage{tabularx}
\usepackage{ragged2e}
\usepackage{lipsum}
\usepackage{longtable}
\usepackage{enumitem}
\usepackage{setspace}
\usepackage{balance}
\pgfplotsset{compat=1.17}
\usepackage{makecell}

\newif\ifshowcomment
\showcommenttrue
\ifshowcomment
\newcommand{\luyao}[1]{\textcolor{blue}{[luyao] #1}}

\else
\newcommand{\luyao}[1]{}
\fi

\begin{document}

\title{Monotonicity for AI ethics and society: \\ An empirical study of the monotonic neural additive model in criminology, education, health care, and finance}


\author[Dangxing Chen]{Dangxing Chen}
\authornote{Contacts: Dangxing Chen: dangxing.chen@dukekunshan.edu.cn, Zu Chongzhi Center for Mathematics and Computational Sciences; Luyao Zhang: luyao.zhang@dukekunshan.edu.cn, Data Science Research Center and Social Science Division; Address: Duke Kunshan University, Duke Avenue No.8, Kunshan, Suzhou, Jiangsu, China, 215316}
 \affiliation{%
 \department{Zu Chongzhi Center for Mathematics and Computational Sciences}
  \institution{Duke Kunshan University}
  \streetaddress{Duke Avenue No.8, Kunshan}
  \city{Suzhou}
  \state{Jiangsu}
  \country{China}
  \postcode{215316}
 }
 \email{dangxing.chen@dukekunshan.edu.cn}

\author[Luyao Zhang]{Luyao Zhang}
\authornotemark[1]
\orcid{0000-0002-1183-2254}
 \affiliation{%
  \department{Data Science Research Center and Social Science Division}
  \institution{Duke Kunshan University}
  \streetaddress{Duke Avenue No.8, Kunshan}
  \city{Suzhou}
  \state{Jiangsu}
  \country{China}
  \postcode{215316}
 }
\email{lz183@duke.edu}

\renewcommand{\shortauthors}{D. Chen and L. Zhang}

\begin{abstract}
\begin{quote}
Act in such a way that you treat humanity, whether in your own person or in the person of any other, never merely as a means to an end, but always at the same time as an end.\\---Immanuel Kant, Grounding for the Metaphysics of Morals
\end{quote}
Algorithm fairness in the application of artificial intelligence (AI) is essential for a better society. As the foundational axiom of social mechanisms, fairness consists of multiple facets. Although the machine learning (ML) community has focused on intersectionality as a matter of statistical parity, especially in discrimination issues, an emerging body of literature addresses another facet---monotonicity. Based on domain expertise, monotonicity plays a vital role in numerous fairness-related areas, where violations could misguide human decisions and lead to disastrous consequences. In this paper, we first systematically evaluate the significance of applying monotonic neural additive models (MNAMs), which use a fairness-aware ML algorithm to enforce both individual and pairwise monotonicity principles, for the fairness of AI ethics and society. We have found, through a hybrid method of theoretical reasoning, simulation, and extensive empirical analysis, that considering monotonicity axioms is essential in all areas of fairness, including criminology, education,  health care, and finance. Our research contributes to the interdisciplinary research at the interface of AI ethics, explainable AI  (XAI), and human-computer interactions (HCIs). By evidencing the catastrophic consequences if monotonicity is not met, we address the significance of monotonicity requirements in AI applications. Furthermore, we demonstrate that MNAMs are an effective fairness-aware ML approach by imposing monotonicity restrictions integrating human intelligence.

\end{abstract}

\ccsdesc[500]{Human-centered computing~Empirical studies in HCI}
\ccsdesc[500]{Applied computing~Education}
\ccsdesc[500]{Applied computing~Law}
\ccsdesc[500]{Applied computing~Economics}
\ccsdesc[500]{Applied computing~Sociology}

\keywords{monotonicity, AI ethics, fairness, human-computer interactions, explainable AI, interpretability, criminology, education, health care, finance}



\maketitle
\section{Introduction}
In recent years, artificial intelligence (AI) and machine learning (ML) have made great strides in a variety of fields by providing highly accurate models for prediction. Meanwhile, public concern has increased over the misuse of ML methods, particularly in high-stakes industries such as criminology, education, health care, and finance. In these areas, accuracy is not the only concern. Blindly relying on machine learning models could result in catastrophic consequences such as unfair decisions. A historic first step toward filling the regulatory gap has been taken by the European Commission with its proposed Artificial Intelligence Act (AIA) \cite{EU2021act}. Furthermore, a review article \cite{carlo2021AI} explains why regulators are required to ensure that AI methods are transparent, explainable, and fair. As a result, we aim to develop regulated ML models that can benefit society in a meaningful way. 
\par
As the foundational axiom of social mechanisms, fairness consists of multiple facets~\cite{sep-social-procedures}. The ML literature has focused on intersectional fairness as a matter of statistical parity: if the probability of the outcome is approximately the same across all subgroups defined by different combinations of protected characteristics, such as race and gender, an AI algorithm is intersectionally fair \cite{foulds2020intersectional,hardt2016equality,zemel2013learning,feldman2015certifying}. Other emerging literature has addressed the necessity to also consider monotonic fairness~\cite{liu2020certified,milani2016fast,you2017deep,sivaraman2020counterexample}. Based on the domain knowledge, many features should exhibit monotonic behavior. In credit scoring, for instance, more past-due debts should result in a lower score. In addition to individual monotonicity, \citet{chen2022monotonic} introduced the monotonic neural additive models (MNAMs), a fairness-aware advancement of the  neural additive model (NAMs) where pairwise monotonicity is also considered. As an example, if a debt is past due, it should result in a lower score if it occurred more recently than in the past. The literature has, unfortunately, neglected pairwise monotonicity. However, a systematic evaluation of MNAM's impact on ML applications is still lacking. In this paper,  we evaluate the significance of applying MNAMs for the fairness of AI ethics and society. Specifically, we answer the following research questions.
\begin{itemize}
    \item {\em Discrepancy in theory.} What are the mathematical and statistical properties that might cause the discrepancy between monotonicity in feature variables and nonmonotonicity in observed data?  
    \item {\em Evidence in applications.} Can MNAMs be used to improve the individual and pairwise monotonicity in ML models, and how do they compare with NAMs in criminology, education, health care, and finance, the four major applications of fairness-aware ML?  
\end{itemize}

We have found, through a hybrid method of theoretical reasoning, simulation, and extensive empirical analysis, that considering individual monotonicity as well as pairwise monotonicity is essential in all areas of fairness, including criminology, education, health care, and finance. First, via a mathematical and statistical deep dive into the monotonicity axioms, we identify the violations of monotonicity as a result of diminishing marginal effects (DMEs) together with relatively large random perturbations in the data-generating process. Since the DME is a commonly acknowledged principle and phenomenon in both the production and consumption theory in economics~\cite{harris2007diminishing,easterlin2005diminishing}, we envision the violations to be widely spread in reality. Then, we present the result of two simulations that verify our conjecture. Finally, we further evaluate the performance of MNAMs by comprehensive empirical analyses of datasets from criminology, education, health care, and finance. We find that using MNAMs significantly reduces the violations of monotonicity axioms compared with other alternatives, such as using NAMs and fully connected neural networks (FCNNs). By evidencing the catastrophic consequences if monotonicity is not met, we address the significance of monotonicity requirements in AI applications. Furthermore, we demonstrate that MNAMs are an effective fairness-aware ML approach by imposing monotonicity restrictions integrating human intelligence. 

The structure of this paper is organized as follows. \Cref{sec:related literature} discusses our contribution to related literature. \Cref{prerequisites} reviews the neural additive model, individual and pairwise monotonicity, and monotonic neural additive models. We analyze types of pairwise monotonicity and causes of violations for monotonicity in \Cref{sec: types and causes of monotonicity}. \Cref{sec: empirical examples} conducts empirical experiments. We conclude and discuss future research in \Cref{sec: conclusion and future research}.

\subsection{Related Literature}
\label{sec:related literature}
Our research contributes to the interdisciplinary research at the interface of AI ethics, axplainable AI (XAI), and human-computer interactions (HCIs). 
\subsubsection{AI ethics}
AI ethics~\cite{sep-ethics-ai} involves the study of foundational questions about the use of AI systems, including issues such as privacy, data manipulation, opacity, bias, employment, morality, etc. Fairness is an important objective of AI ethics~\cite{madaio2020co,pessach2022review}. 
Fairness-aware machine learning~\cite{bird2019fairness} aims to develop fairness-enhancing mechanisms for machine learning algorithms. Our research contributes to the literature by analyzing the importance of pairwise monotonicity and evidencing MNAMs as a new fairness-aware machine learning solution. 
\newline
\par
Our research also shows how the synthesis of literature could yield further advancements. For example, through the use of explainable models, we are able to gain a better understanding of how the models work so as to further identify potential fairness issues. Moreover, by incorporating human insights, we can bring more humanity to ML models by enforcing social norms and ethical concerns. 

\subsubsection{Explainable AI (XAI)}
Explainable artificial intelligence (XAI)~\cite{xu2019explainable} is an emerging research addressing the black-box problem of an AI system including machine learning algorithms. \citet{gunning2019xai} define XAI as a system that is intelligible to humans by providing explanations such as monotonicity with respect to certain variables and correlated variables obeying particular relationships. \citet{gade2019explainable} further elaborate on the importance of explainability in applying AI to assist in decision-making in the industry. Furthermore, a variety of approaches have been proposed to disentangle the obscurity of ML models, including \citet{agarwal2021neural},\citet{yang2020enhancing}, \citet{yang2021gami}, \citet{chen2018interpretable}, and~\citet{dumitrescu2022machine}. Our research contributes to the XAI literature by demonstrating how the MNAMs can be used to improve the explainability of the previous NAMs in applying machine learning to assist predictions in criminology, education, health care, and finance. 

\subsubsection{Human-Computer Interactions (HCI)}
Among the literature on human-computer interaction (HCI)~\cite{amershi2019guidelines}, \citet{dellermann2021future} elaborate on the necessity of hybrid intelligence systems, combining human and artificial intelligence, in real-world business applications to collectively achieve superior and self-optimizing decisions. \citet{krishna2022socially} show that deep learning AI agents perform better by learning from human interactions in computer vision and natural language processing tasks. Our research contributes to the literature by demonstrating how the integration of human insights on monotonicity can improve the performance of machine learning algorithms, especially for biased datasets with limited observations in part of the data domains.

\section{Prerequisites}
\label{prerequisites}
Assume we have $\mathcal{D} \times \mathcal{Y}$, where $\mathcal{D}$ is the dataset with $n$ samples and $p$ features and $\mathcal{Y}$ is the corresponding numerical values in regression and labels in classification. We assume the data-generating process (DGP) is
\begin{align} \label{eq:regre}
y = f(\*x) + \epsilon
\end{align}
for regression problems and 
\begin{align}
 y|\*x = \text{Bernoulli}(f(\*x))    
\end{align}
for binary classification problems. Then machine learning (ML) methods are applied to approximate $f$.



\subsection{Neural additive models}
Fully-connected neural networks (FCNNs) have been very successful for high-dimensional complex functions, due to their universal approximation property \cite{cybenko1989approximation,hornik1991approximation,kubat1999neural,hassoun1995fundamentals}. Despite their success in approximation, their complicated deep layers with massively connected neurons prevent us from interpreting the result. Neural additive models (NAMs) \cite{agarwal2021neural} improve the explainability of FCNNs by restricting the architecture of neural networks (NNs). NAMs belong to the family of generalized additive models (GAMs) of the form
\begin{align*}
g(\mathbb{E}[y|\*x]) = \beta + f_1(x_1) + \dots + f_p(x_p),
\end{align*}
where $\*x = (x_1, \dots, x_p)$ is the input with $p$ features, $y$ is the target variable, $g(\cdot)$ is the link function, and is chosen as the logistic link function $g(z) = \frac{z}{1-z}$ in our study. For NAMs, each $f_i$ is parametrized by an NN.



\subsection{Individual and pairwise monotonicity}

Individual and pairwise monotonicity are both common in many applications. Without loss of generality, assume that all monotonic constraints are increasing. Suppose $\~{\alpha}$ is the list of all monotonic features and $\neg \~{\alpha}$ its complement, then the input $\*x$ can be partitioned into $\*x = (\*x_{\~{\alpha}}, \*x_{\neg \alpha})$. Suppose $\mathcal{X}$, $\mathcal{X}_{\~{\alpha}}$, $\mathcal{X}_{\neg \~{\alpha}}$ are spaces of $\*x, \*x_{\~{\alpha}}, \*x_{\neg \~{\alpha}}$, respectively. Then we have the following definition of individual monotonicity. 
\begin{definition}
We say $f$ is individually monotonic with respect to $\*x_{\~{\alpha}}$ if
\begin{align} \label{eq:mono_con1}
f(\*x_{\~{\alpha}}, \*x_{\neg \~{\alpha}}) \leq f(\*x'_{\~{\alpha}}, \*x_{\neg \~{\alpha}}), \forall \*x_{\~{\alpha}} \leq \*x'_{\~{\alpha}},  \forall \*x_{\~{\alpha}}, \*x_{\~{\alpha}}' \in \mathcal{X}_{\alpha}, \forall \*x_{\neg \~{\alpha}} \in \mathcal{X}_{\neg \~{\alpha}},
\end{align}
where $\*x_{\~{\alpha}} \leq \*x_{\~{\alpha}}'$ denotes the inequality for all entries, i.e., $x_{\alpha_i} \leq x_{\alpha_i}'$.
\end{definition}

Analogous to \eqref{eq:mono_con1}, we partition $\*x = (x_{\beta},x_{\gamma},\*x_{\neg})$, where $x_{\beta}, x_{\gamma} \in \mathcal{X}_{\beta,\gamma}$ and $\*x_{\neg} \in \mathcal{X}_{\neg}$. Here, $x_{\beta}$ and $x_{\gamma}$ are in the same space so that they can be compared. Then we have the following definition of pairwise monotonicity.
\begin{definition}
We say $f$ is monotonic with respect to $x_{\beta}$ over $x_{\gamma}$ if 
\begin{align}\label{eq:mono_con2}
    f(x_{\beta},x_{\gamma}+c,\*x_{\neg}) \leq f(x_{\beta}+c,x_{\gamma},\*x_{\neg}),  x_{\beta} = x_{\gamma}, \forall x_{\beta}, x_{\gamma} \in \mathcal{X}_{\beta,\gamma}, \forall \*x_{\neg} \in \mathcal{X}_{\neg}, \forall c \in \mathbb{R}^+. 
\end{align}
\end{definition}

\subsection{Monotonic neural additive models}
Monotonic neural additive models (MNAMs) \cite{chen2022monotonic} enforce both individual and pairwise monotonicity in NAMs. Suppose $\~{\alpha}$ is the list containing all individual monotonic features. Suppose $\*u$ and $\*v$ are lists for pairwise monotonic features such that $f$ is monotonic with respect to $u_i$ over $v_i$. In MNAMs, monotonicities are verified as follows:
\begin{itemize}
\item Monotonicity of individual features:
\begin{align} \label{eq:mono1}
\min_i \ \min_x \ \frac{\partial f_{\alpha_i}(x)}{\partial x}  \geq 0.
\end{align}
\item Monotonicity of pairwise features:
\begin{align} \label{eq:mono2}
\min_i \ \min_x \ \left(\frac{\partial f_{u_i}(x)}{\partial x} - \frac{\partial f_{v_i}(x)}{\partial x}\right) \geq 0.
\end{align}
\end{itemize}
The parameters are optimized by 
\begin{align} \label{eq:MNAM_loss}
\min_{\~{\Theta}} \ \ell(\~{\Theta}) + \lambda h_1(\~{\Theta}) + \eta h_2(\~{\Theta}),
\end{align}
where $\ell$ is the mean-squared error for regressions and log-likelihood functions for classifications, and
\begin{align*}
h_1(\~{\Theta}) &= \sum_j \sum_{i} \max \left(0, -\frac{\partial f_{\alpha_i} (x_j; \Theta_{\alpha_i})}{\partial x}   \right)^2, \\
h_2(\~{\Theta}) &=  \sum_j \sum_{i} \max \left( 0, -\frac{\partial f_{u_i} (x_j;\Theta_{u_i})}{\partial x} + \frac{\partial f_{v_i} (x_j;\Theta_{v_i})}{\partial x} \right)^2,
\end{align*}
provided that $f_i$ is sufficiently smooth. 
In the optimization procedure, $\max(0,\cdot)$is replaced with $\max(\epsilon,\cdot)$.

The algorithm gradually increases $\lambda$ and $\eta$ until the penalty terms vanish. The two-step procedure is summarized in Algorithm~\ref{alg:MNAM}. The NAM satisfies all required monotonic constraints \eqref{eq:mono1} and \eqref{eq:mono2} and is referred to as the monotonic neural additive model (MNAM). 


\begin{algorithm}[h]
\caption{Monotonic additive model}\label{alg:MNAM}
\begin{algorithmic}[1]
\State \textbf{Initialization}: $\lambda=0$ and $\eta=0$
\State Train an NAM by \eqref{eq:MNAM_loss}
\While{$\min(h_1,h_2)>0$}
\State Increase $\lambda$ if $h_1>0$ and increase $\eta$ if $h_2>0$
\State Retrain the NAM by \eqref{eq:MNAM_loss}.
\EndWhile
\end{algorithmic}
\end{algorithm}

\section{Types and causes of monotonicity}
\label{sec: types and causes of monotonicity}



\subsection{Types of pairwise monotonicity in practice}

The concept of individual monotonicity has been extensively studied. Our focus here is on pairwise monotonicity. There is a common occurrence of pairwise monotonicity in practice, but it is often overlooked. The two most common types are as follows. 

\begin{itemize}
\item Pairwise monotonicity of time. It is common for features to contain information from different time periods. Consider credit scoring. Assume $x_i$ represents a count of the number of past due payments within a year and $x_j$ represents a count of the overall number of past payments one year ago. A past due payment should affect our output more heavily if it occurred recently. This restriction is necessary for two reasons: 
\begin{enumerate}
\item As a general rule, it is expected that the importance of information will diminish over time under normal circumstances.
\item More importantly, even if the pattern of data suggests the opposite of what we are trying to enforce, we should still enforce it for fairness' sake. As forgiveness is an important aspect of our moral lives, we should not place more emphasis on past mistakes than on present mistakes. By providing people with opportunities to improve themselves, they may be able to behave better. In fact, this is consistent with many policies in practice; for example, a delinquency record can be removed approximately seven years after it occurred. When building models, we must take into account such consequences, which are not captured in machine learning models, into account. 
\end{enumerate}
\item Pairwise monotonicity of severity. There are certain events that are intrinsically more important than others. When studying recidivism, we could use two features to track the prior felonies and misdemeanors, and felonies tend to be more serious crimes involving violence. We must incorporate this restriction for two reasons:  
\begin{enumerate}
\item As a general rule, a more severe event serves as a strong indicator of future behavior. 
\item It is important to have a fair reward and punishment system. A violation of such fairness could potentially misguide people's behavior in the future. If a person finds that the punishment for a misdemeanor is greater than that of a felony, he may commit the misdemeanor into a felony next time. A consequence such as this cannot be seen in the accuracy setting at the moment, but we must avoid it. 
\end{enumerate}
\end{itemize}

\subsection{Violations resulting from diminishing marginal effects and data distributions}

While monotonicity should be satisfied in theory, empirical data do not always reveal the perfect pattern. Consider the regression framework \eqref{eq:regre}. In practice, we are unable to observe the $f(x_i)$ due to the high-dimensionality of $\*x$, but it is possible to observe the marginal conditional expectation $\mathbb{E}[y|x_i]$. Suppose $\frac{\partial f}{\partial x_i} (\*x) > 0$. Then we have $\frac{\partial}{\partial x_i} \mathbb{E}[y|x_i] = \mathbb{E} \left[\frac{\partial f(\*x)}{\partial x_i} \middle| x_i \right] > 0$, which means that $\mathbb{E}[y|x_i]$ should also be monotonically increasing. Similarly, suppose $f$ is monotonic with respect to $x_i$ over $x_j$, then, we would have $\frac{\partial}{\partial x_i} \mathbb{E}[y|x_i] > \frac{\partial}{\partial x_j} \mathbb{E}[y|x_j]$. In practice, we could replace the expectation $\mathbb{E}[y|\cdot]$ with the average $\overline{y}|\cdot$ and visualize monotonicity. Empirical data, however, often violate such monotonies as a result of estimation errors. Furthermore, we find that diminishing marginal effects as well as the distribution of data enhances violations.

There is a common phenomenon in the nature of diminishing marginal effects. That is, the increment decreases with the increment of the given feature. The following definition can be defined mathematically to describe this phenomenon. 
\begin{definition}[Diminishing marginal effect]
Assume $f$ is differentiable; then, we say that $x_i$ has a diminishing marginal effect (DME) if 
\begin{align*}
\frac{\partial f}{\partial x_i}(\*x) \geq 0, å\frac{\partial^2 f}{\partial x_i^2}(\*x) \leq 0, \forall \*x \in \mathcal{X}.
\end{align*}
\end{definition}
 Although the DME presents an interesting social phenomenon for the study of, it poses a challenge to the training of machine learning models. As function values are quite similar when the increment is small, ML models can easily ignore monotonicity due to random perturbations.

The distribution of features is another cause of the empirical violation of monotonicity. Specifically, a distribution does not always follow a uniform pattern in practice. Tails are a common occurrence. It is expected that ML models train well with respect to the behavior of the dataset in the dense region; however, confidence will be lost in the sparse region due to a lack of samples. It is often the case that we do not have sufficient samples at the tails of monotonic features. It is particularly common for monotonic features to be associated with punishments, such as the number of past due payments in a credit scoring or the number of crimes in the recidivism rate. When these circumstances arise, it is natural for there to be a smaller amount of data in the right tail.

The training of ML models could be misguided by these violations. The result will be a reduction in the accuracy of unseen data as well as an increase in unfairness. Therefore, we must strictly enforce monotonic requirements for ML models. 

\subsection{Simulation}
Using simulation studies, we demonstrate how monotonicity is violated in data. Our discussion of DMEs will be based on log utility functions, which are commonly used by economists and financiers. The distribution of data is assumed to follow the Poisson distribution. 

\subsubsection{Individual monotonicity}

Let us consider a function $f: [0,\infty) \rightarrow [0,\infty)$ 
\begin{align*}
f(x) = \alpha \log(c+x),
\end{align*}
where $c \geq 1$. $f(x)$ is a monotonic function in the domain. Suppose $X \sim \text{Poisson}(\lambda)$ and $\epsilon \sim \mathcal{N}(0,\sigma^2)$. In theory, the expectation on $y$ should be monotonic
\begin{align*}
\frac{\partial}{\partial x} \mathbb{E}[y|x] = \frac{\alpha}{c+x}>0.
\end{align*}
However, in practice, estimations may be affected by DMEs, noise, and the distribution of data. A DME's impact is measured by looking at different regions of $f$: as $c+x$ increases, $f$ becomes flatter. Magnitudes of $\sigma$ reflect noise levels. The impact of a distribution is evaluated by considering different values of $\lambda$: a smaller $\lambda$ indicates a distribution that is closer to the origin. Our goal is to investigate the impact of parameters on the results. It is, however, difficult to analyze their relationship analytically. As a result, simulation studies are conducted for a variety of parameters. We generate 10,000 samples each time. We then repeat the experiment 1,000 times and calculate the probability that monotonicity is violated empirically for $\overline{y}|x$, where $\overline{y}$ calculates the average of $y$. For simplicity, we focus on the calculation for $x \in [0,4]$. A summary of the results can be found in Table~\ref{tab:simu_indi}. The results indicate that all parameters have significant impacts on violations, suggesting that empirical violations are quite common.

\begin{table}[h]
    \centering
    \caption{Simulation results for the empirical violation of individual monotonicity. Violations are affected significantly by diminishing marginal effects, data distributions, and noise levels.  }
    \vspace{5mm} 
    \begin{tabular}{cccccc}
    \hline
    DME & $\alpha=1$ & $\sigma = 0.2$ & $\lambda=0.5$ & \\ \hline
    $c$  & $5$ & $10$ & $15$ & 20  \\ \hline
    Ratio of violations & $1.8\%$ & $10.4\%$ & $15.8\%$ & $21.6\%$ \\ \hline 
    Noise & $\alpha=1$ & $c = 10$ & $\lambda=0.5$ & \\ \hline
    $\sigma$  & $0.1$ & $0.2$ & $0.3$ & 0.4  \\ \hline
    Ratio of violations & $0.7\%$ & $8.3\%$ & $20.1\%$ & $24.9\%$ \\ \hline 
    Distribution & $\alpha=1$ & $c = 10$ & $\sigma=0.2$ & \\ \hline
    $\lambda$  & $0.3$ & $0.4$ & $0.5$ & 0.6  \\ \hline
    Ratio of violations & $29.8\%$ & $20.1\%$ & $7.9\%$ & $3.4\%$ \\ \hline 
    \end{tabular}
    \label{tab:simu_indi}
\end{table}

\subsubsection{Pairwise monotonicity}

We now generalize the result to 2-dimensions to consider pairwise monotonicity. 
Let us consider the following  $f: \mathbb{R}_+^2 \cup (0,0) \rightarrow [0,\infty)$
\begin{align}
f(x_1,x_2) = \alpha \log(c+x_1) + \beta \log(c+x_2).
\end{align}
where $c \geq 1$ and $\alpha>\beta$. Therefore, $f$ has the DME and is monotonic with respect to $x_1$ over $x_2$. Suppose the $X_1$ and $X_2$ are generated by Poisson distributions with parameters $\lambda$ and $\mu$. Theoretically, marginal monotonicities could be observed:
\begin{align*}
\frac{\partial}{\partial x_1} \mathbb{E}[y|x_1] & = \frac{\alpha}{c+x_1}> 0, \\
\frac{\partial}{\partial x_2} \mathbb{E}[y|x_2] & = \frac{\beta}{c+x_2}>0, \\
\frac{\partial}{\partial x_1} \mathbb{E}[y|x_1] &>\frac{\partial}{\partial x_2} \mathbb{E}[y|x_2], \ \forall x_1=x_2 \geq 0.
\end{align*}
In regard to pairwise monotonicity, there are too many parameters to consider. The results are based on one set of parameters for the sake of simplicity. We let $c=10$, $\alpha=1.2$, $\beta=1$, $\lambda=0.5$, $\mu=0.4$, $\sigma = 0.2$, and the rest of the setup is the same as before. Both monotonies are checked empirically for violations. The result is striking: there are $5.4\%$ violations of individual monotonicity for $x_1$, $21.6\%$ violations of individual monotonicity for $x_2$, and $69.2\%$ violations of pairwise monotonicity. According to this result, violations of pairwise monotonicity are more common than violations of individual monotonicity. In this case, we are comparing two functions at their tails, in which estimations can be quite noisy due to diminishing marginal effects and data distribution.





\section{Empirical examples}
\label{sec: empirical examples}

This section evaluates the performance of models for a variety of datasets, including four examples in different fields, namely, criminology, education,  health care, and finance.  Specifically, we compare fully-connected neural networks (FCNNs), neural additive models (NAMs), and monotonic neural additive models (MNAMs). For simplicity, we randomly split datasets into training ($80\%$) and testing ($20\%$). The classification error, area under the curve (AUC), and confusion matrices are provided to measure performance. Then, we provide a detailed postanalysis for fairness.

As part of the model architecture, we use the same structure for FCNNs and individual neural networks (NNs) in NAMs and MNAMs so that they have the same number of parameters. NNs contain 1 hidden layer with 2 units, logistic activation, and no regulation.  Since such simple architectures are compared favorably with results from the literature, we do not explore more complex architectures.


In this study, for simplicity, we have neglected other advanced ML methods, such as random forest (RF) and extreme gradient boosting (XGBoost). For some datasets, a comparison of FCNNs with these methods can be found in \cite{agarwal2021neural,yang2021gami}.  We shall highlight that the focus of our method is on transparency, interpretability, and fairness, rather than accuracy. Furthermore, we exclude features in the protected group, such as race and gender, but we do not put further restrictions on optimization as it is not the focus of this paper. However, we want to note that this part can be easily performed.

\subsection{COMPAS}

\subsubsection{Data description}
COMPAS, which is used to guide bail, sentencing, and parole decisions is a proprietary score developed to predict recidivism risk. It has been criticized for racial bias\cite{COMPAS2016crime,dressel2018accuracy,tan2018distill}. A report published by ProPublica in 2016 provides recidivism data for defendants in Broward County, Florida \cite{Propublica2016COMPAS}. We focus on the simplified cleaned dataset provided in \cite{dressel2018accuracy}.~\footnote{\url{https://farid.berkeley.edu/downloads/publications/scienceadvances17/}} Three thousand and fifty-one ($45\%$) of the 7,214 observed defendants committed a crime within two years. This study uses a binary response variable, recidivism, as the response variable. The dataset here contains nine features, which were selected after some feature selection was conducted. 
\begin{itemize}
\item $x_1$: races include White (Caucasian), Black (African American), Hispanic, Asian, Native American, and Others
\item $x_2$: sex, male or female
\item $x_3$: age
\item $x_4$: total number of juvenile felony criminal charges
\item $x_5$: total number of juvenile misdemeanor criminal charges
\item $x_6$: total number of nonjuvenile criminal charges
\item $x_7$: a numeric value corresponding to the specific criminal charge
\item $x_8$: an indicator of the degree of the charge: misdemeanor or felony
\item $x_9$: a numeric value between 1 and 10 corresponding to the recidivism risk score generated by COMPAS software (a small number corresponds to a low risk, and a larger number corresponds to a high risk).
\item $y$: whether the defendant recidivated two years after the previous charge
\end{itemize}
To avoid discrimination, we further exclude race and sex. The COMPAS score is also excluded because it is not the focus of this study and is correlated with other features, making its interpretation more difficult. As there are too few samples, we truncate the number of juveniles exceeding three. Otherwise, if monotonicity is requested, NN functions will become flat, which is not useful. To ensure individual monotonicity, we place restrictions on the number of juvenile felony and misdemeanor charges, prior nonjuvenile charges, and indicators of the degree of the charge. Pairwise monotonicity requires that our prediction be monotonic with respect to felony offenses over misdemeanor offenses. There is a pairwise monotonicity of severity in this case. Felonies are usually more serious crimes involving violence and carry heavier punishments. The imposition of this pairwise monotonicity is consistent with the criminal justice system. 

\subsubsection{Results}
\begin{figure}[h]
\includegraphics[scale=0.5]{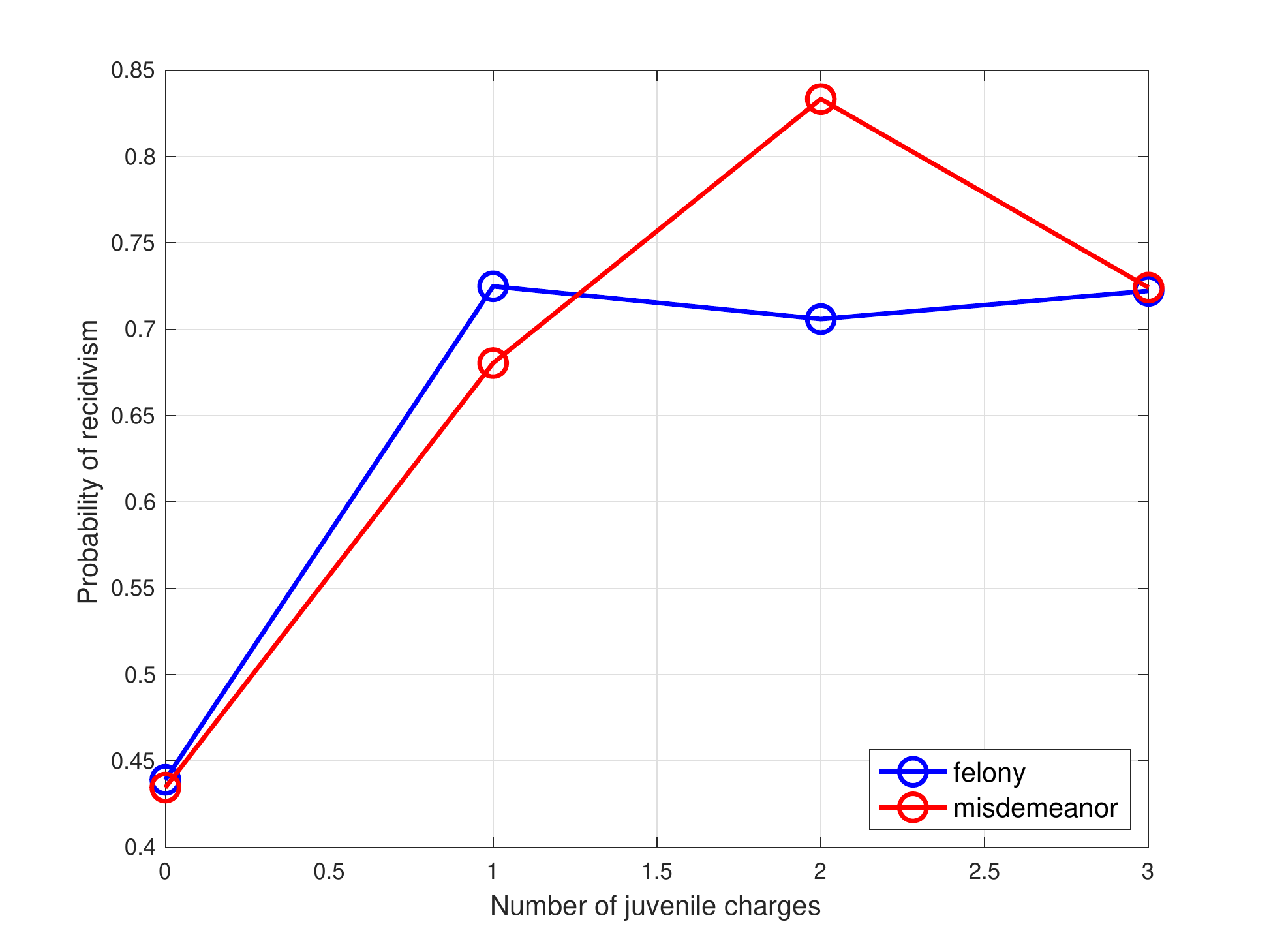}
\caption{The marginal probability of recidivism as a function of the number of juvenile misdemeanors. Monotonicity is violated both individually and pairwise. }
\label{fig:recid_juv}
\end{figure}

\begin{figure}[h]
\includegraphics[scale=0.5]{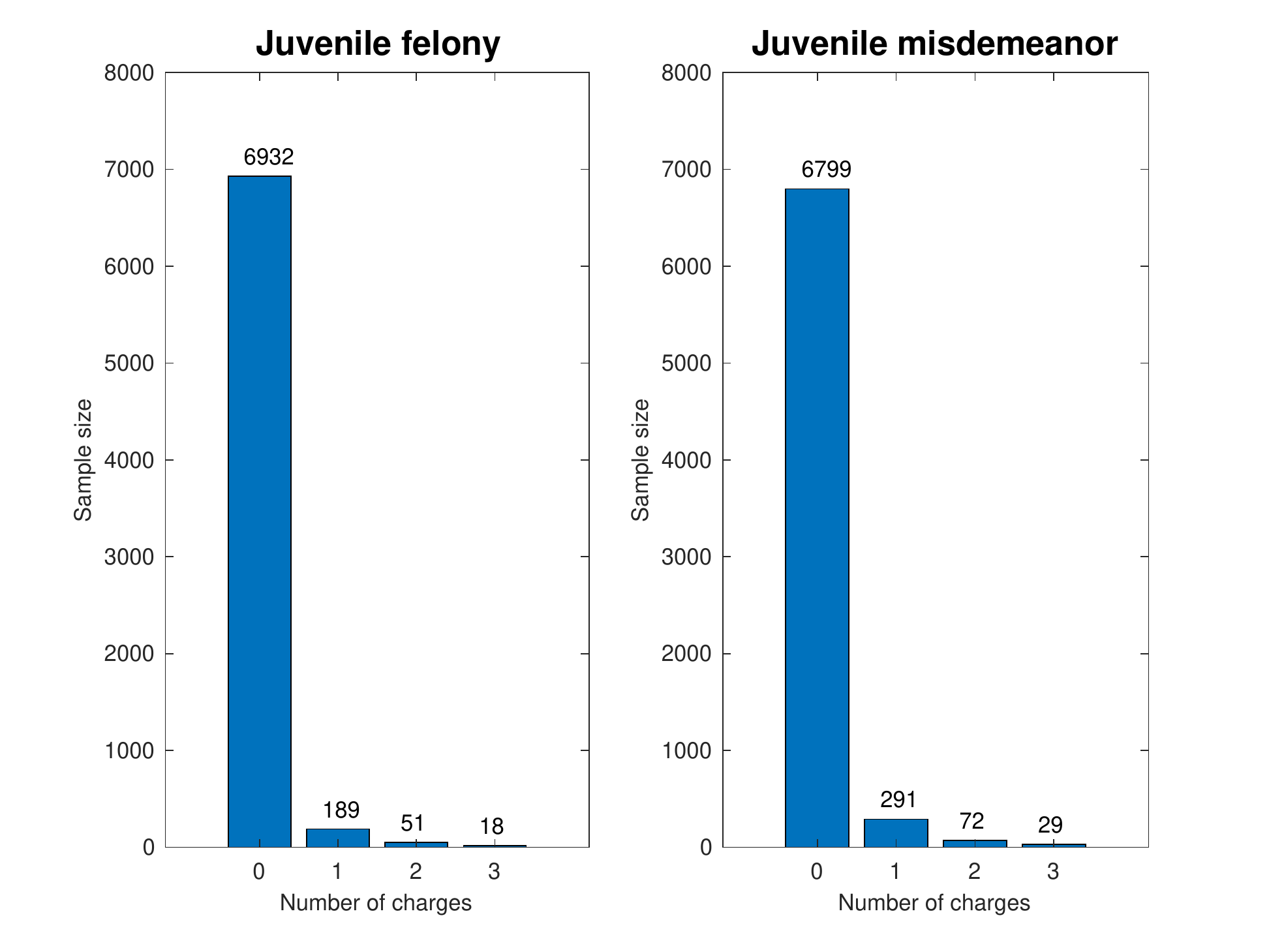}
\caption{The histograms represent the number of charges. For a greater number of charges, there are fewer samples.}
\label{fig:charge_hist}
\end{figure}

We first observe that monotonicities could be violated in empirical data. A plot showing the marginal probability of recidivism based on the number of juvenile felonies and misdemeanors can be found in Figure~\ref{fig:recid_juv}. It has been observed that individual monotonicities are violated, especially in misdemeanor cases. Between two and three misdemeanors, there is a sharp decline.  In addition to violations of individual monotonicity, violations of pairwise monotonicity are also observed in Figure~\ref{fig:recid_juv}. For misdemeanors, the probability of recidivism is greater than for felonies when there are two charges. As analyzed in Section~\ref{sec: types and causes of monotonicity}, one potential cause is due to diminishing marginal effects, as shown in Figure~\ref{fig:recid_juv}. There is a significant increment between zero and one felony charge, however, further increments are not clearly visible. The second potential cause is the rapid decay of the distribution tail. A truncated histogram of the number of charges is shown in Figure~\ref{fig:charge_hist}. In the majority of samples, there are no prior charges, and then the distribution rapidly declines. The number of samples can be considered too small to be statistically significant for more than two crimes. This may explain the empirical violations observed in Figure~\ref{fig:recid_juv}. The probability of more than two charges is somewhat close as a result of the DME. With limited data, estimations of probability become unreliable for more than one charge for both felony and misdemeanor offenses. Such a pattern could misguide machine learning training without proper instructions.

We then apply NN, NAM, and MNAM to the dataset. The results are summarized in Table~\ref{tab:COMPAS_result}. The same level of accuracy is achieved using the family of neural additive models such as the FCNN. Based on these results, the MNAM and NAM should be preferred over FCNN in terms of their transparency and explainability. 



\begin{table}[h]
\caption{A comparison of model performance on the COMPAS dataset. The accuracy of the three methods is similar.}
\parbox{.45\linewidth}{
\centering
    \begin{tabular}{ccc}
    \hline
    Model/Metrics & Classification error  & AUC  \\ \hline
    FCNN & 33.7$\%$ & $71.5\%$ \\ \hline
    NAM  & 34.0$\%$ & $71.5\%$ \\ \hline
    MNAM  & 33.4$\%$ & $71.7\%$ \\ \hline
    \end{tabular}
}
\hfill
\parbox{.5\linewidth}{
\centering
    \begin{tabular}{ccc}
    \hline
    FCNN  & Predicted: Yes & Predicted: No  \\ \hline
    Actual: Yes  & 472 & 335  \\ \hline
    Actual: No  & 272 & 724  \\ \hline
    NAM  & Predicted: Yes & Predicted: No  \\ \hline
    Actual: Yes & 466  & 341  \\ \hline
    Actual: No  & 272 & 724  \\ \hline
    MNAM  & Predicted: Yes & Predicted: No  \\ \hline
    Actual: Yes  & 453  & 354  \\ \hline
    Actual: No  & 248 & 748  \\ \hline
    \end{tabular}
}
\label{tab:COMPAS_result}
\end{table}

\begin{figure}[h]
\centering
\includegraphics[scale=0.5]{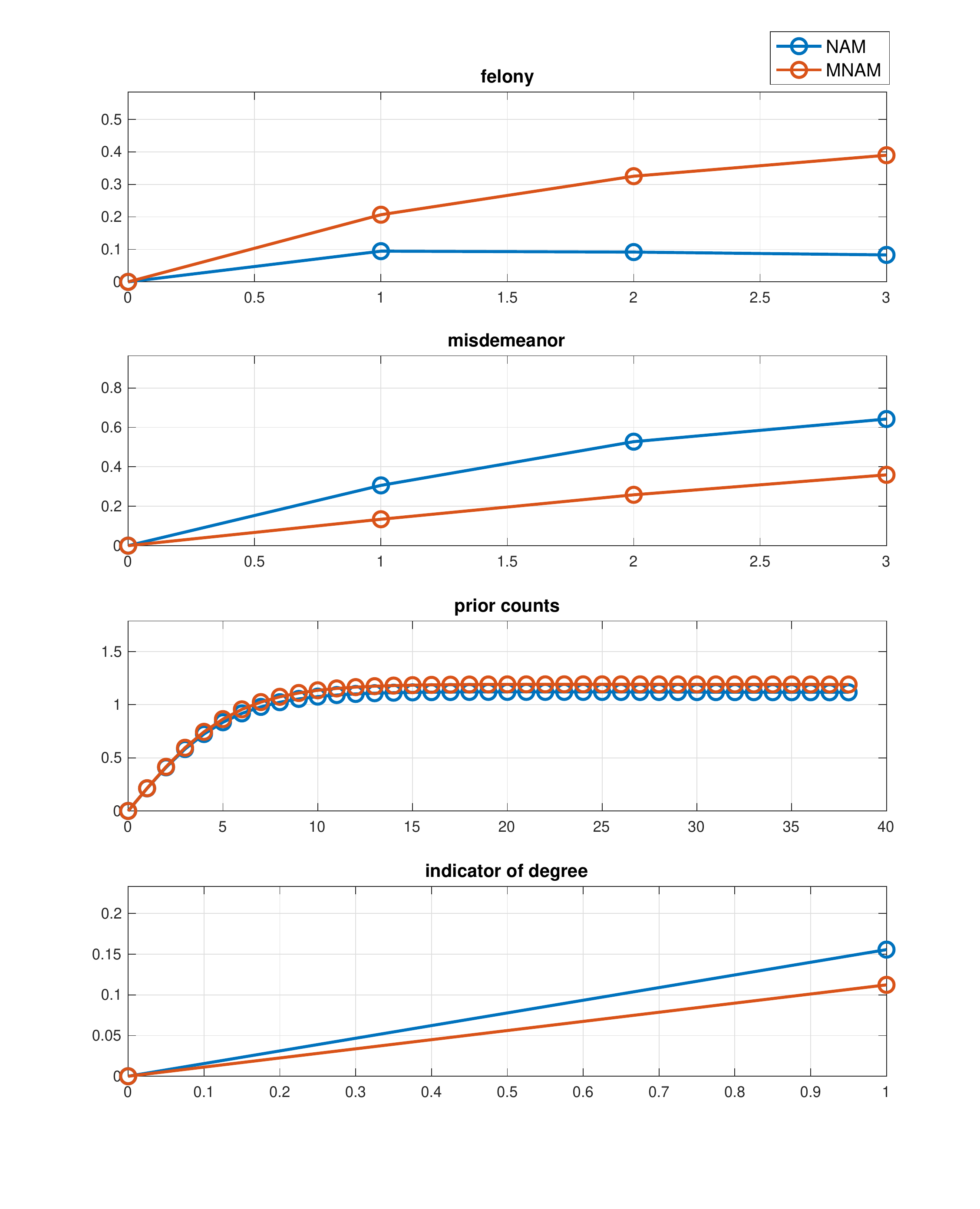}
\caption{A comparison of the behaviors of the NAM and MNAM for individual monotonic features in the COMPAS dataset. For the feature felony, the individual monotonicity is slightly violated in the NAM. }
\label{fig:COMAS_fn}
\end{figure}

\begin{figure}[h]
\centering
\includegraphics[scale=0.5]{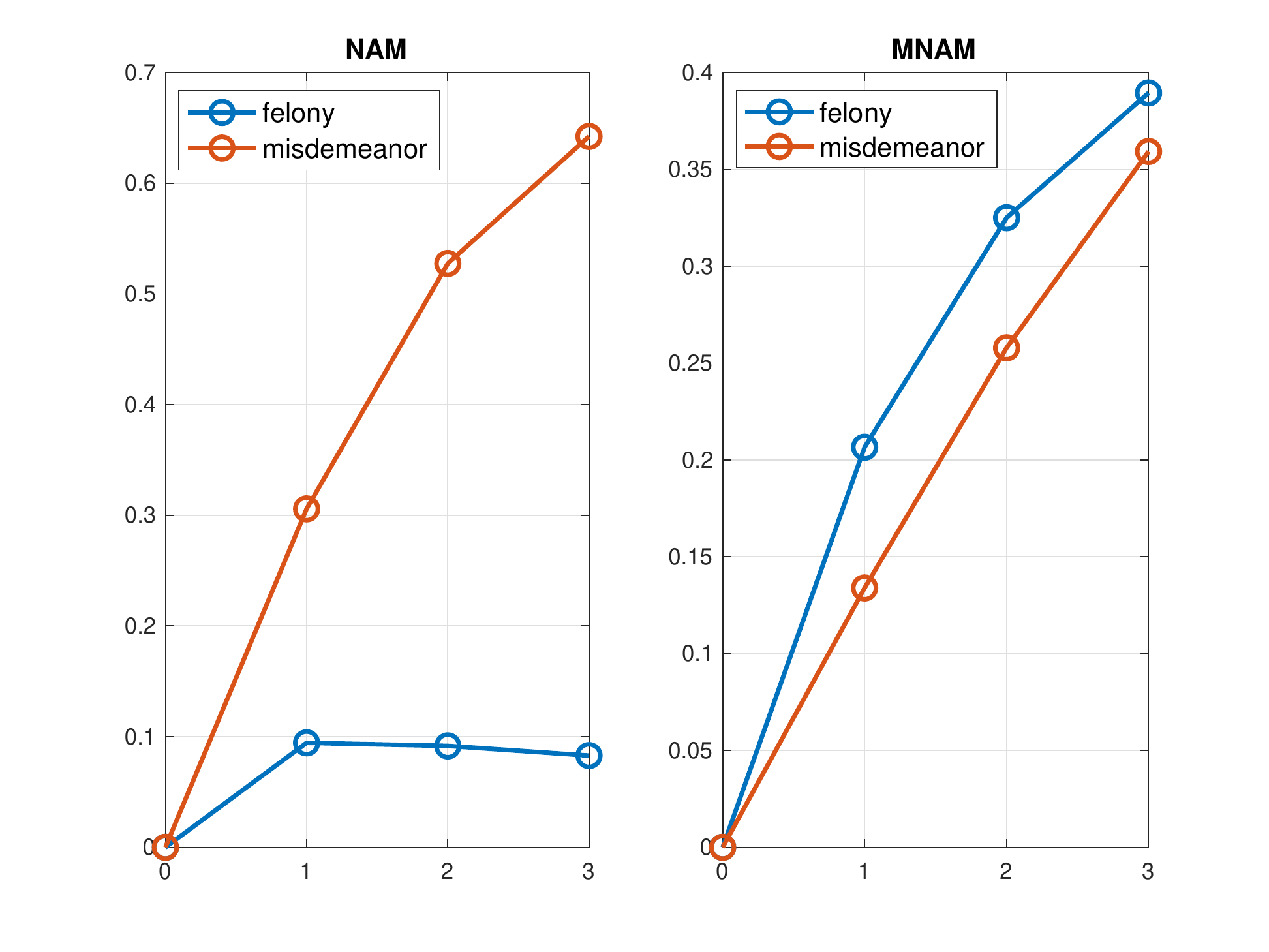}
\caption{A comparison of the behaviors of pairwise monotonic features on the COMPAS dataset when using NAM and MNAM. The NAM violates pairwise monotonicity. }
\label{fig:COMAS_pair}
\end{figure}

We then check the algorithmic fairness. The monotonic features of this dataset are analyzed. We begin by analyzing the monotonicity of individual features in Figure~\ref{fig:COMAS_fn}. This training result does not appear to be affected by a serious violation of individual monotonicity by the NAM. However, if we examine the felony in more detail, we can observe a slow decrease after one charge by the NAM, which is corrected in the MNAM. Then we check the pairwise monotonicity. The results are somewhat shocking, as shown in Figure~\ref{fig:COMAS_pair}. In the NAM, the concept of fairness is violated by weighing misdemeanors much more heavily than felonies. In the MNAM, on the other hand, the impacts are reweighted without compromising accuracy. individual and pairwise monotonicity are enforced in the MNAM to send a clear message to criminals that crimes are taken seriously, and that felonies are more serious than misdemeanors. Take a moment to consider your actions before taking them. 

\subsection{Law school dataset}

\subsubsection{Data description}

The law school dataset \cite{wightman1998lsac}~\footnote{\url{https://github.com/tailequy/fairness_dataset/tree/main/Law_school}} concerns information on the probability of passing the bar examination. In 1991, 163 law schools in the United States were surveyed by the Law School Admission Council (LSAC).  From the total of 18,692 observations, 16,856 ($90\%$) people passed the bar for the first time. If, for instance, universities wish to award scholarships based on the likelihood of passing the bar examination, fairness could be important. In this study, the response variable is a binary variable, pass. There are 11 features in this dataset. 
\begin{itemize}
\item $x_1$: the student’s decile in the school given his grades in Year 1
\item $x_2$: the student’s decile in the school given his grades in Year 3
\item $x_3$: the student’s LSAT score
\item $x_4$: the student’s undergraduate GPA
\item $x_5$: the student's first year law school GPA
\item $x_6$: the student's cumulative law school GPA
\item $x_7$: whether the student will work full-time or part-time
\item $x_8$: the student’s family income bracket
\item $x_9$: whether the student is a male or female
\item $x_{10}$: tier
\item $x_{11}$: race
\item $y$: whether the student passed the bar exam on the first try
\end{itemize}
Race and sex were excluded. The law school GPA (LGPA) is calculated on different scales for the first year and the cumulative. To make a comparison, we scale them. For all grade-related features ($x_1-x_6$), we require individual monotonicity, as well as pairwise monotonicity for $x_2$ over $x_1$ and $x_6$ over $x_5$. In the latter cases, the requirement indicates the pairwise monotonicity of time: the more recent information should be regarded as more valuable.

\subsubsection{Results}

NN, NAM, and MNAM are applied to the dataset. A summary of the results can be found in Table~\ref{tab:law_result}. The performance of all models is similar. 



\begin{table}[h]
\caption{A comparison of model performance on a law school dataset. The accuracy of the three methods is similar. }
\parbox{.45\linewidth}{
\centering
    \begin{tabular}{ccc}
    \hline
    Model/Metrics  & Classification error & AUC  \\ \hline
    FCNN & $8.9\%$ & $88.9\%$ \\ \hline
    NAM & $8.9\%$ & $88.9\%$ \\ \hline
    MNAM & $8.8\%$ & $88.9\%$ \\ \hline
    \end{tabular}
}
\hfill
\parbox{.5\linewidth}{
\centering
    \begin{tabular}{ccc}
    \hline
    FCNN  & Predicted: Yes & Predicted: No  \\ \hline
    Actual: Yes  & 4155 & 60  \\ \hline
    Actual: No  & 358 & 100  \\ \hline
    NAM  & Predicted: Yes & Predicted: No  \\ \hline
    Actual: Yes  & 4147  & 68  \\ \hline
    Actual: No  & 349 & 109  \\ \hline
    MNAM  & Predicted: Yes & Predicted: No  \\ \hline
    Actual: Yes  & 4144  & 71  \\ \hline
    Actual: No  & 342 & 116  \\ \hline
    \end{tabular}
}
\label{tab:law_result}
\end{table}

\begin{figure}[h]
\centering
\includegraphics[scale=0.5]{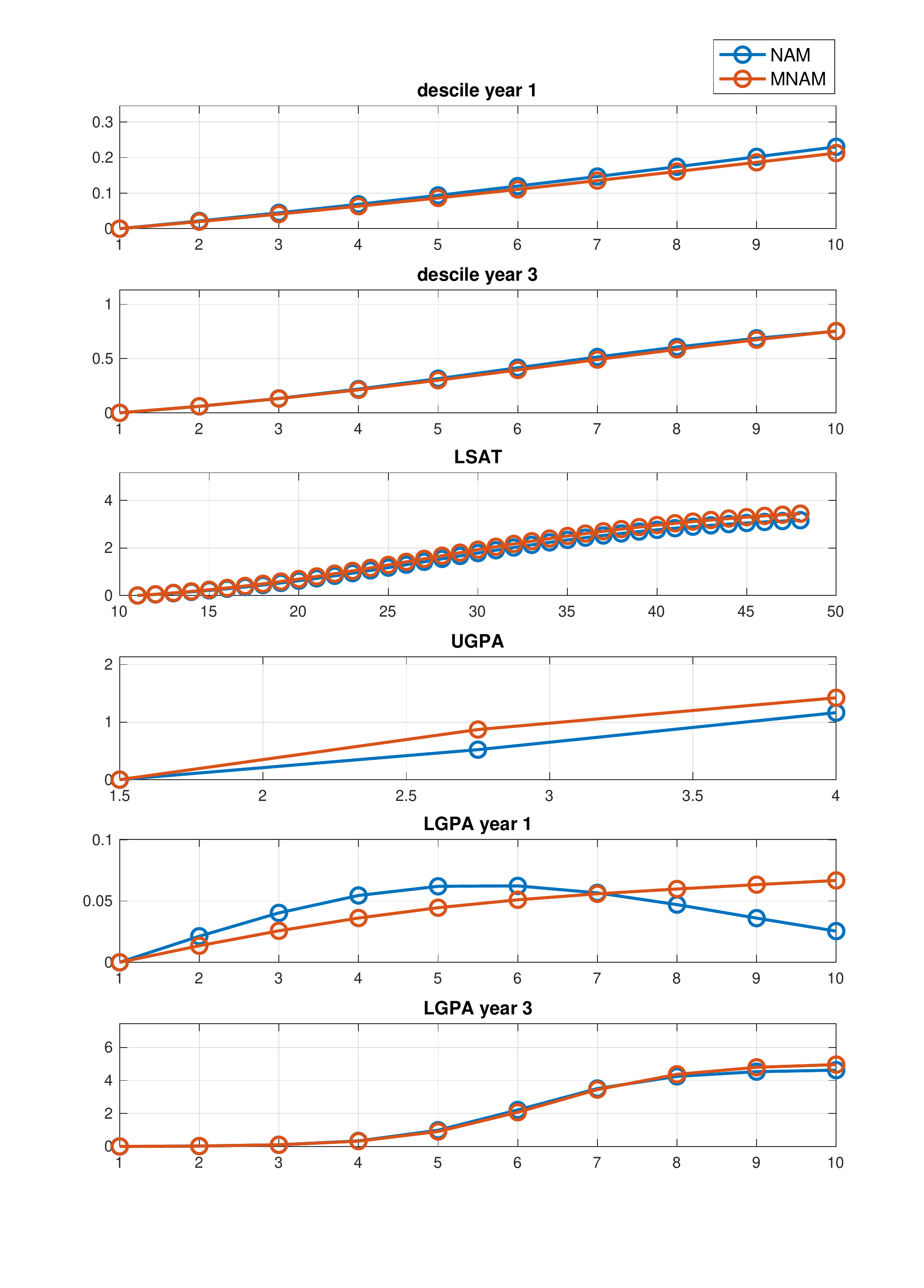}
\caption{A comparison of the behavior of individual monotonic features using NAM and MNAM on the law school dataset. The NAM violates individual monotonicity for LGPA at year 1. }
\label{fig:law_fn}
\end{figure}

\begin{figure}[h]
\centering
\begin{subfigure}{.5\textwidth}
  \centering
  \includegraphics[width=1.0\linewidth]{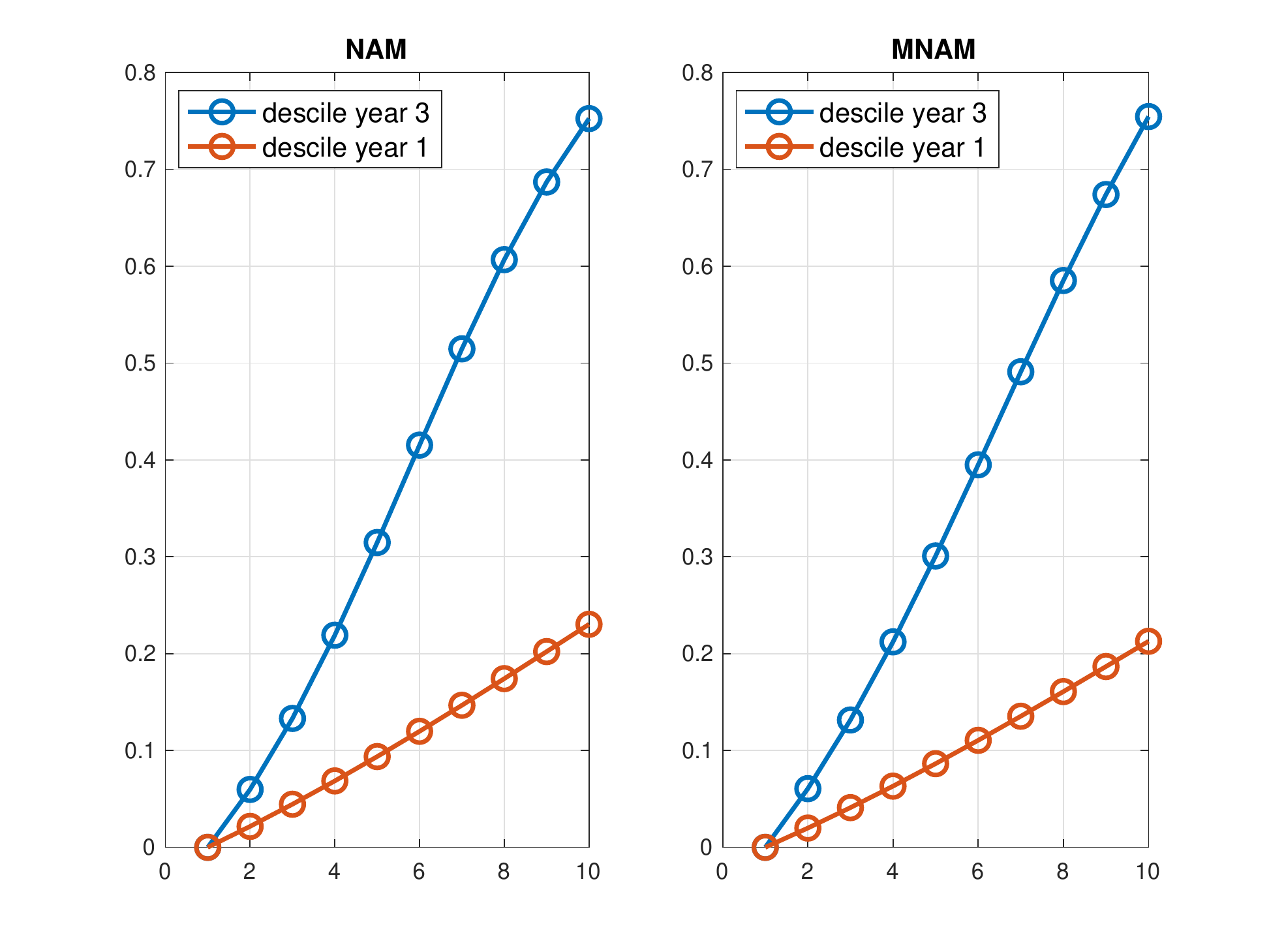}
\end{subfigure}%
\begin{subfigure}{.5\textwidth}
  \centering
  \includegraphics[width=1.0\linewidth]{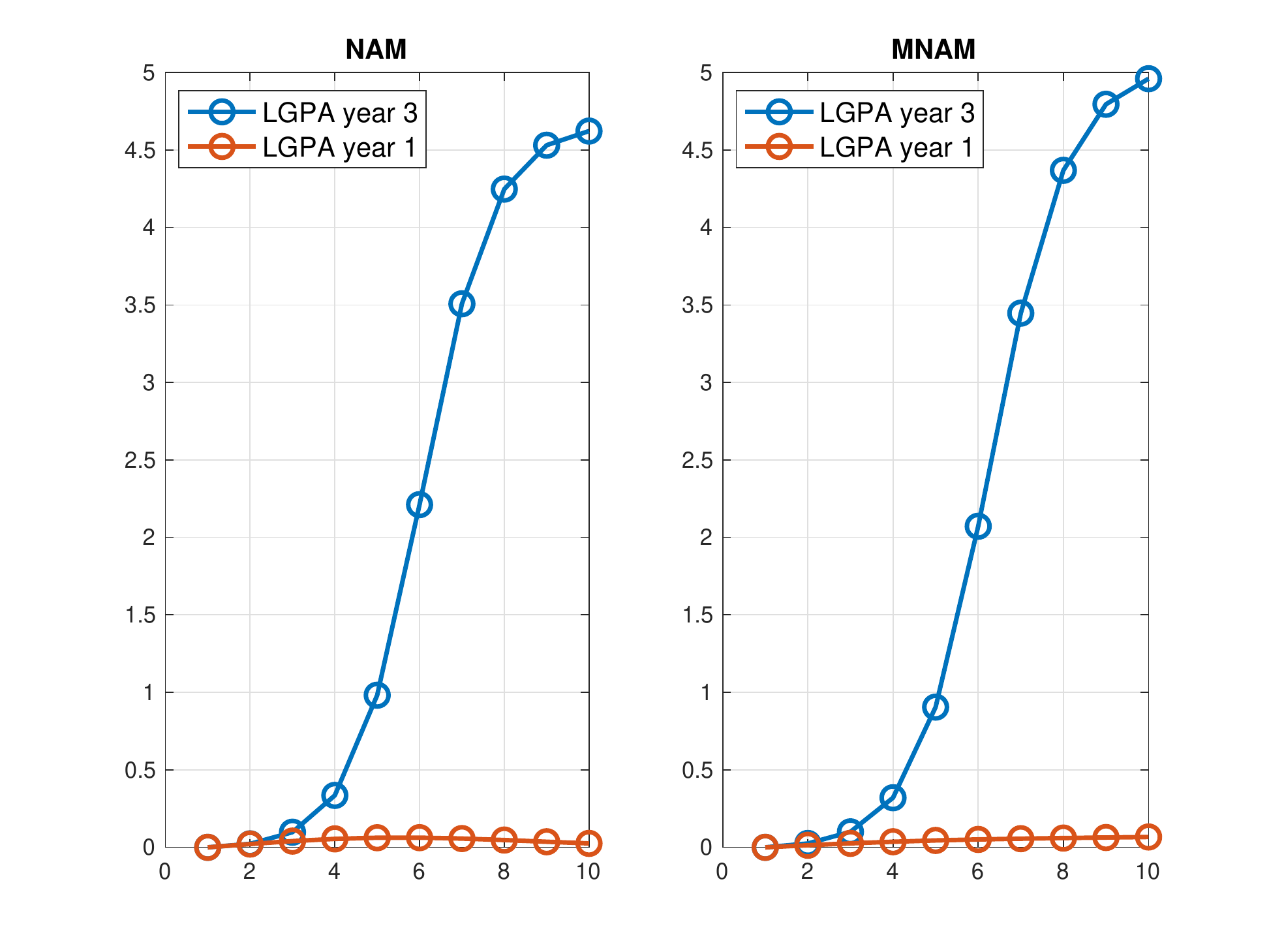}
\end{subfigure}
\caption{Comparison of the behavior for pairwise monotonic features using NAM and MNAM on the law school dataset.}
\label{fig:law_pair}
\end{figure}

After that, we check the algorithmic fairness. First, we analyze the monotonicity of individual features in Figure~\ref{fig:law_fn}. In this training result, the LGPA's behavior in year 1 is problematic when using NAM. The overall magnitude of its impact is small, resulting in a misleading pattern. Due to its small magnitude, it does not have a significant impact on accuracy. However, it causes unfairness to students who did well in their first year of law school. If this NAM is used, it may discourage students from concentrating on their studies during their first year of law school. There is no doubt that this is not the message the school wishes to convey to students. In contrast, students who have performed well during their first year of study are recognized in the MNAM. As a result of pairwise monotonicity in Figure~\ref{fig:law_pair}, violations are not observed in the NAM. 

\subsection{Thoracic surgery dataset}

\subsubsection{Data description}

The thoracic surgery dataset \cite{zikeba2014boosted}~\footnote{\url{https://archive.ics.uci.edu/ml/datasets/Thoracic+Surgery+Data}} contains information about the post-operative life expectancy of lung cancer patients. Wroclaw Thoracic Surgery Centre collected retrospective data on patients who underwent major lung resections for primary lung cancer between 2007 and 2011.  Out of the 470 observations, 70 ($15\%$) people died within one year of their surgery. Fairness could be important here, for example, if doctors must decide which operation will be carried out first based on the condition of the patient. This study uses a binary response variable, death, as its response variable. A total of 16 features are included in this dataset. 
\begin{itemize}
\item $x_1$: diagnosis---specific combination of ICD-10 codes for primary and secondary as well as multiple tumours, if any
\item $x_2$: forced vital capacity 
\item $x_3$: volume that has been exhaled at the end of the first second of forced expiration
\item $x_4$: performance status - Zubrod scale
\item $x_5$: pain before surgery (T,F)
\item $x_6$: hemoptysis before surgery (T,F)
\item $x_7$: dyspnea before surgery (T,F)
\item $x_8$: cough before surgery (T,F)
\item $x_9$: weakness before surgery (T,F)
\item $x_{10}$: T in clinical TNM - size of the original tumor, from OC11 (smallest) to OC14 (largest)
\item $x_{11}$: Type 2 DM - diabetes mellitus (T,F)
\item $x_{12}$: MI up to 6 months (T,F)
\item $x_{13}$: peripheral arterial diseases (T,F)
\item $x_{14}$: smoker (T,F)
\item $x_{15}$: asthma (T,F)
\item $x_{16}$: age at surgery 
\item $y$: 1-year survival period - (T) true value if died (T,F)
\end{itemize}
There are clearly many monotonic relationships among features. Despite this, we are not confident in the ability to place all possible restrictions since we are not experts in this field. As an example, we focus on three features to illustrate the performance. Nevertheless, experts are encouraged to add additional restrictions, which is not difficult. In this case, we believe that the probability of death should monotonically increase with respect to hemoptysis and dyspnea over coughing, as hemoptysis involves coughing up blood and dyspnea involves difficulty breathing. 

\subsubsection{Results}
NN, NAM, and MNAM are applied to the dataset. The results are summarized in Table~\ref{tab:surgery_result}. In general, the performances of the two models are not very different, although only nondeath events tend to be predicted when using FCNN. Other methods may be available to improve certainty accuracy measurements. Nevertheless, we do not elaborate on these topics here since they are not the main subject of the paper. Our primary goal is to demonstrate the potential unfairness caused by ML models. 



\begin{figure}[h]
\centering
\begin{subfigure}{.5\textwidth}
  \centering
  \includegraphics[width=1.0\linewidth]{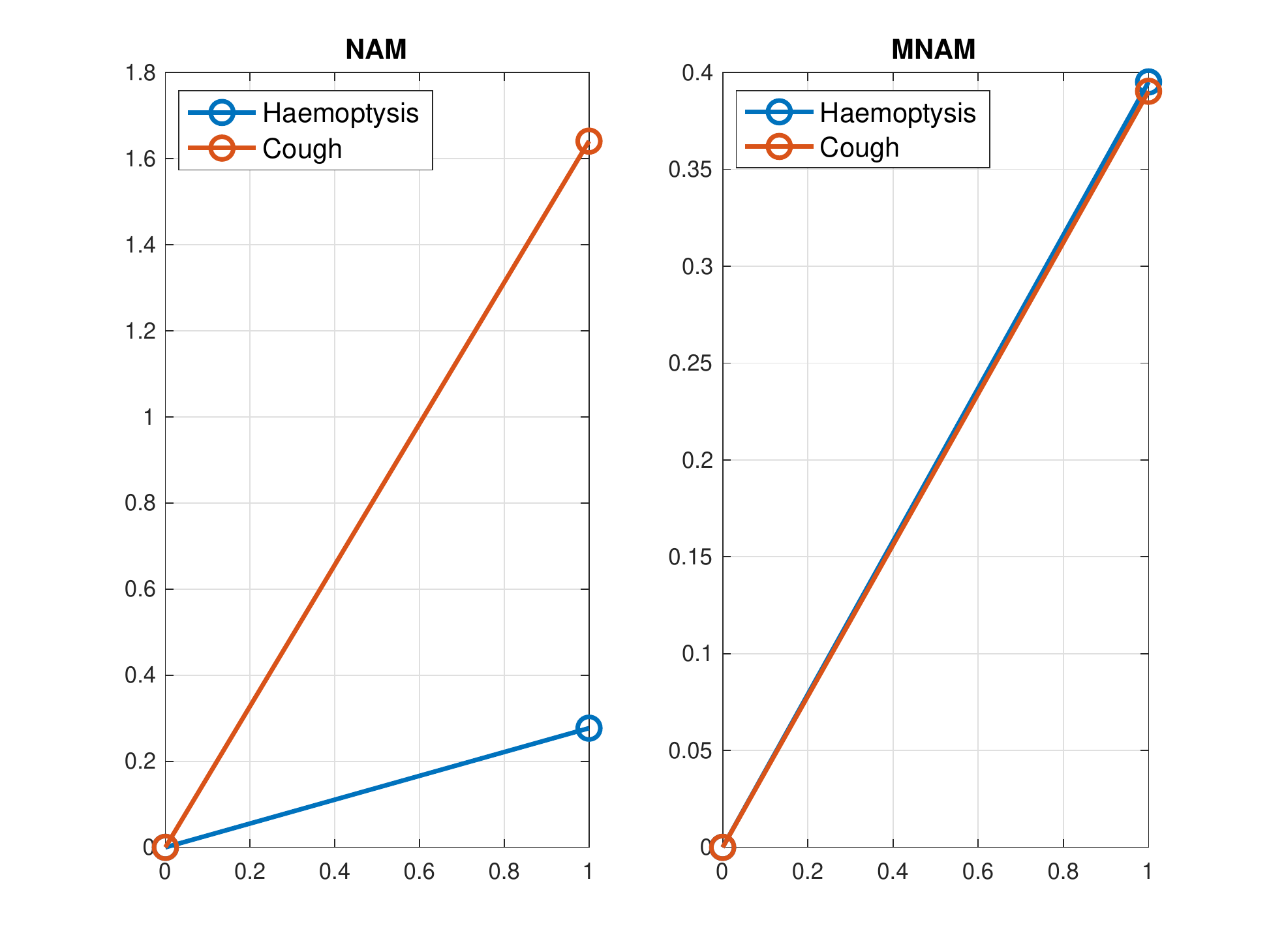}
\end{subfigure}%
\begin{subfigure}{.5\textwidth}
  \centering
  \includegraphics[width=1.0\linewidth]{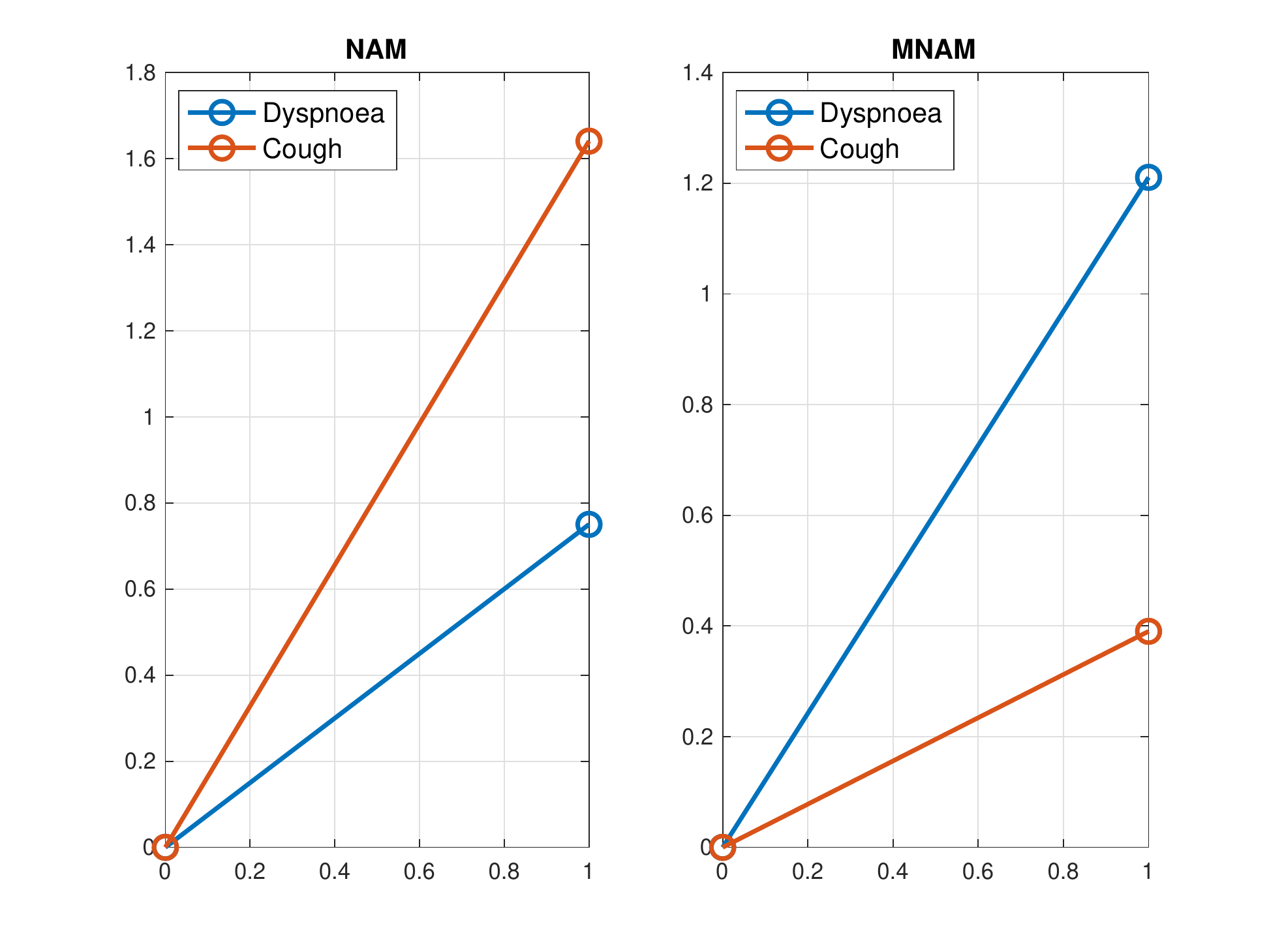}
\end{subfigure}
\caption{A comparison of the behavior of pairwise monotonic features using NAM and MNAM on the thoracic surgery dataset. Both cases violate pairwise monotonicity in the NAM.}
\label{fig:surgery_pair}
\end{figure}

\begin{table}
\caption{A comparison of model performance on the thoracic dataset. }
\parbox{.45\linewidth}{
\centering
    \begin{tabular}{ccc}
    \hline
    Model/Metrics  & Classification error & AUC  \\ \hline
    FCNN & $13.7\%$ & $64.8\%$ \\ \hline
    NAM & $20.5\%$ & $64.8\%$ \\ \hline
    MNAM & $20.5\%$ & $66.2\%$ \\ \hline
    \end{tabular}
}
\hfill
\parbox{.45\linewidth}{
\centering
    \begin{tabular}{ccc}
    \hline
    FCNN  & Predicted: Yes & Predicted: No  \\ \hline
    Actual: Yes  & 0 & 16  \\ \hline
    Actual: No  & 0 & 101  \\ \hline
    NAM  & Predicted: Yes & Predicted: No  \\ \hline
    Actual: Yes  & 0  & 16  \\ \hline
    Actual: No  & 8 & 93  \\ \hline
    MNAM  & Predicted: Yes & Predicted: No  \\ \hline
    Actual: Yes  & 0  & 16  \\ \hline
    Actual: No  & 8 & 93  \\ \hline
    \end{tabular}
}
\label{tab:surgery_result}
\end{table}
We then verify the fairness of the algorithm. For this training, individual monotonicity is satisfied, so we skip it in order to save space. This result for pairwise monotonicity is striking, as shown in Figure~\ref{fig:surgery_pair}. In the opinion of the NAM, coughing is more dangerous than hemoptysis and dyspnea. Despite the fact that we are not doctors, such indications do not seem to be appropriate. A preference for cough as a stronger indicator will certainly lead to some unfair outcomes for patients. As mentioned in~\cite{agarwal2021neural}, the family of NAMs is not a causal model. We do not know what caused the outcome but rather extracted useful information from it. The imposition of monotonicity, however, could enhance the reasonableness and fairness of NAMs. 

\subsection{FICO credit score dataset}
\subsubsection{Data description}
The FICO credit score dataset \footnote{\url{https://community.fico.com/s/explainable-machine-learning-challenge}}~\footnote{\url{https://github.com/5teffen/FICO-xML-Challenge/tree/master/xML\%20Challenge\%20Dataset\%20and\%20Data\%20Dictionary}} has been used to obtain an explainable machine learning approach for calculating credit scores. In this study, researchers examined an anonymized dataset of Home Equity Line of Credit (HELOC) applications made by real homeowners. It is a line of credit offered by a bank as a percentage of a home's equity. The customers in this dataset have requested a credit line in the range of $\$5,000 - \$150,000$. In essence, the task is to determine whether the applicant will repay their HELOC account within 2 years based on the information in their credit report about them. Using the prediction, the lender decides whether and how much credit to extend to the homeowner. A total of 23 features are included in this dataset.
\begin{itemize}
\item $x_1$: Consolidated version of risk makers
\item $x_2$: Months since oldest trade open
\item $x_3$: Months since most recent trade open
\item $x_4$: average months in file
\item $x_5$: number of satisfactory trades
\item $x_6$: number of trades 60+ ever
\item $x_7$: number of trades 90+ ever
\item $x_8$: percent trades never delinquent
\item $x_9$: months since most recent delinquency
\item $x_{10}$: max delinquency/public records in the last 12 months
\item $x_{11}$: max delinquency ever
\item $x_{12}$: number of total trades (total number of credit accounts)
\item $x_{13}$: number of trades open in last 12 months
\item $x_{14}$: percent install trades
\item $x_{15}$: months since most recent inquiry excluding 7 days
\item $x_{16}$: number of inquiries in last 6 months
\item $x_{17}$: number of inquiries in the last 6 months excluding 7 days
\item $x_{18}$: net fraction revolving burden. This is a revolving balance divided by the credit limit
\item $x_{19}$: net fraction of installment burden. This is the installment balance divided by the original loan amount
\item $x_{20}$: number of revolving trades with balance
\item $x_{21}$: number of installment trades with balance
\item $x_{22}$: number of bank trades with high utilization ratio
\item $x_{23}$: percent of trades with balance
\item $y$: whether the account is in default or not
\end{itemize}
For this dataset, we focus on the monotonicity of the delinquency features. That is, the probability of default should be individually monotonically increasing with respect to maximum delinquency, and furthermore, more recent delinquency should be more important than historical delinquency. 

\subsubsection{Results}

NN, NAM, and MNAM are applied to the dataset. The results are summarized in Table~\ref{tab:Fico_result}. All methods perform similarly for this dataset, which is consistent with the finding in \cite{chen2018interpretable}. 

\begin{figure}[h]
\centering
\includegraphics[scale=0.4]{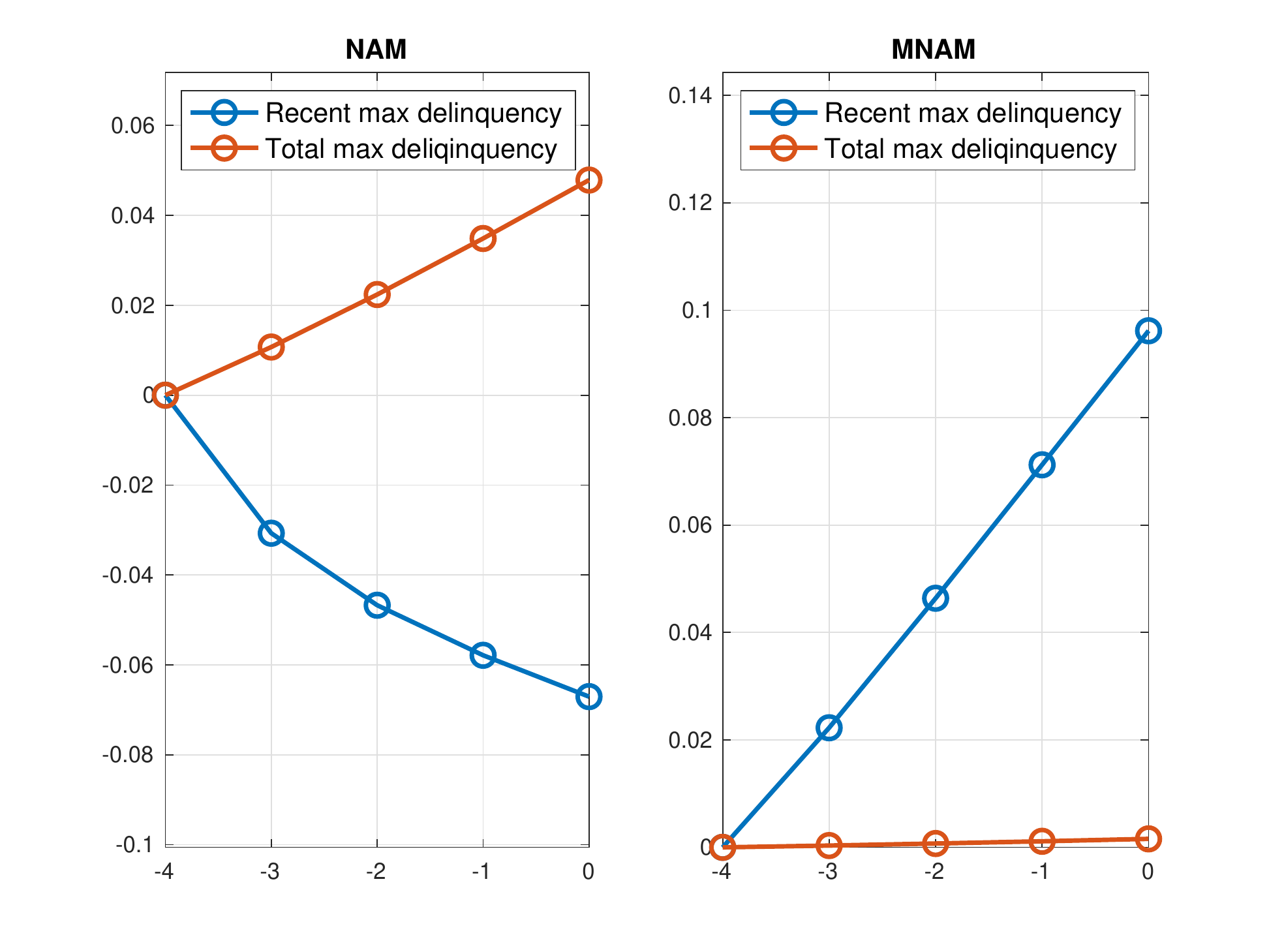}
\caption{Comparison of the behavior of pairwise monotonic features using NAM and MNAM on the FICO credit scoring dataset. In the NAM, pairwise monotonicity is violated.}
\label{fig:Fico_pair}
\end{figure}

\begin{table}
\caption{A comparison of model performance on the FICO dataset. All three methods are equally accurate.}
\parbox{.45\linewidth}{
\centering
    \begin{tabular}{ccc}
    \hline
    Model/Metrics  & Classification error & AUC  \\ \hline
    FCNN & $27.8\%$ & $80.0\%$ \\ \hline
    NAM & $27.3\%$ & $80.0\%$ \\ \hline
    MNAM & $27.0\%$ & $80.1\%$ \\ \hline
    \end{tabular}
}
\hfill
\parbox{.45\linewidth}{
\centering
    \begin{tabular}{ccc}
    \hline
    FCNN  & Predicted: Yes & Predicted: No  \\ \hline
    Actual: Yes  & 838 & 388  \\ \hline
    Actual: No  & 297 & 942  \\ \hline
    NAM  & Predicted: Yes & Predicted: No  \\ \hline
    Actual: Yes  & 859  & 367  \\ \hline
    Actual: No  & 306 & 933  \\ \hline
    MNAM  & Predicted: Yes & Predicted: No  \\ \hline
    Actual: Yes  & 851  & 375  \\ \hline
    Actual: No  & 290 & 949  \\ \hline
    \end{tabular}
}
\label{tab:Fico_result}
\end{table}
We then verify the fairness of the algorithm. We compare the performance of delinquency features using NAM and MNAM, as shown in Figure~\ref{fig:Fico_pair}. Here, features from $x=-4$ to $x=0$ represent 30/60/90/120 days delinquent, and derogatory comments, respectively. When using NAM, both the individual and pairwise monotonicity are violated for the recent delinquency feature. In MNAM, this problem is avoided, and recent features are emphasized. 

\section{Conclusion and Future Research}
\label{sec: conclusion and future research}
This paper examines individual monotonicity and pairwise monotonicity in fairness-related fields, such as criminology, education, health care, and finance. Both types of monotonicities are essential for fairness and two common types of pairwise monotonicity are discussed. Because of the diminishing marginal effect and the rapid decay of the distribution, empirical data may present misleading patterns, which could lead to unfair machine-learning models. Using monotonic neural additive models, we conduct extensive empirical studies in a variety of fields. We demonstrate that monotonicity is often violated if we are not careful, how it could lead to catastrophic outcomes, and how monotonic neural additive models could improve them. Finally, we would like to emphasize the importance of monotonicity when building trustworthy machine-learning models in fairness-sensitive fields, and we would like to encourage researchers to develop features and models that are monotonicity-aware. Our approach to monotonicity-aware AI can be extended in the three facets of research questions, methodology, and application scenarios.
\subsection{Research Question: Discrimination}
Our study shows how MNAMs outperform existing NAMs in explainability and fairness with datasets considering individual and pairwise monotonicity. Future research could explore the interactions of monotonic and intersectional fairness. For instance, \citet{wang2020deontological} find that by enforcing individual monotonicity, the general additive models (GAMs)~\cite{hastie2017generalized}~\footnote{https://github.com/tensorflow/
lattice/blob/master/docs/tutorials} also improve on statistical fairness as a byproduct. We could further investigate the side-effect of pairwise monotonicity on statistical parity and discrimination.

\subsection{Methodology: Causal Model}
In MNAM, the human insights of monotonicity are integrated with the family of NAMs, which is not a causal model~\cite{agarwal2021neural}. However, even causal models depend on various unverified assumptions~\cite{scholkopf2021toward}. Future research could further integrate the domain expertise of monotonicity and pairwise monotonicity to verify the interpretability of causal models. 

\subsection{Application Scenario: AI Economist}
Our results show the effectiveness of MNAMs in criminology, education, health care, and finance. Recently, economists have had a growing interest in integrating AI technology for advising individual, institutional, and policy decisions~\cite{athey2019machine}. For example, \citet{kleinberg2018human} shows that machine learning can improve judges' decisions in criminology if integrated into an economic framework. Future research can further incorporate pairwise monotonicity to study economic decisions including market design~\cite{milgrom2018artificial}, corporate governance~\cite{hilb2020toward}, and macroeconomic policies~\cite{goulet2022machine}.

\bibliographystyle{ACM-Reference-Format}
\bibliography{monotonicity}

\end{document}